\documentclass[12pt,doublespacing]{spieman}  % 12pt font required by SPIE;
\usepackage{amsmath,amsfonts,amssymb}
\usepackage{graphicx}
\usepackage{setspace}
\usepackage{tocloft}
\usepackage{multirow}
\usepackage{lipsum}
\usepackage{rotating}

\title{Exploring the significance of using perceptually relevant image decolorization method for scene classification}

\author[1] {V. Sowmya}
\author[2] {D. Govind}
\author [3]{ K. P. Soman}

\affil{Centre for Computational Engineering and Networking (CEN)\\ Amrita School of Engineering, Coimbatore \\ Amrita Vishwa Vidyapeetham, Amrita University, India\\
v\_sowmya@cb.amrita.edu, d\_govind@cb.amrita.edu, kp\_soman@amrita.edu}

\renewcommand{\cftdotsep}{\cftnodots}
\begin{document}
\maketitle

\begin{abstract}

A color image contains luminance and chrominance components representing the intensity and color information respectively. The objective of the work presented in this paper is to show the significance of incorporating the chrominance information for the task of scene classification. An improved color-to-grayscale image conversion algorithm by effectively incorporating the chrominance information is proposed using color-to-gay structure similarity index (C2G-SSIM) and singular value decomposition (SVD) to improve the perceptual quality of the converted grayscale images. The experimental result analysis based on the image quality assessment for image decolorization called C2G-SSIM and success rate (Cadik and COLOR250 datasets) shows that the proposed image decolorization technique performs better than 8 existing benchmark algorithms for image decolorization. In the second part of the paper, the effectiveness of incorporating the chrominance component in scene classification task is demonstrated using the deep belief network (DBN) based image classification system developed using dense scale invariant feature transform (SIFT) as features. The levels of chrominance information incorporated by the proposed image decolorization technique is confirmed by the improvement in the overall scene classification accuracy . Also, the overall scene classification performance is improved by the combination of models obtained using the proposed and the conventional decolorization methods.
\end{abstract}

\footnote{This article must be cited as: Sowmya Viswanathan, Govind Divakaran, Kutti Padanyl Soman, "Significance of perceptually relevant image decolorization for scene classification," J. Electron. Imaging 26(6), 063019 (2017), doi: 10.1117/1.JEI.26.6.063019. }

\section{Introduction}
The two main modules of any image classification system are feature extraction and classification. The features are extracted from the input color image in RGB space by processing the three color planes independently. However, most of the existing image classification systems, in particular, scene classification systems converts the input color image to grayscale, (so called color-to-grayscale image conversion) prior to feature extraction \cite{OT2001, Xie2012, Lazebnik2006, Mandar, Fei-Fei2005}.

The color-to-grayscale image conversion is performed to reduce the computational complexity of the features extracted from the single gray plane rather than from the three dimensional color planes. Also, the incorporation of significant chrominance information along with the contrast details in the converted grayscale images are essential, as the features required to categorize or classify the scene information are extracted from the converted grayscale images. But, the major challenge in the color-to-grayscale image conversion is the preservation of luminance information along with the chrominance contrast in the converted grayscale images as in the original color image \cite{Alsam2009}.

\begin{figure*} [htbp]
\centering
\includegraphics[width=\textwidth,height=2.25in]{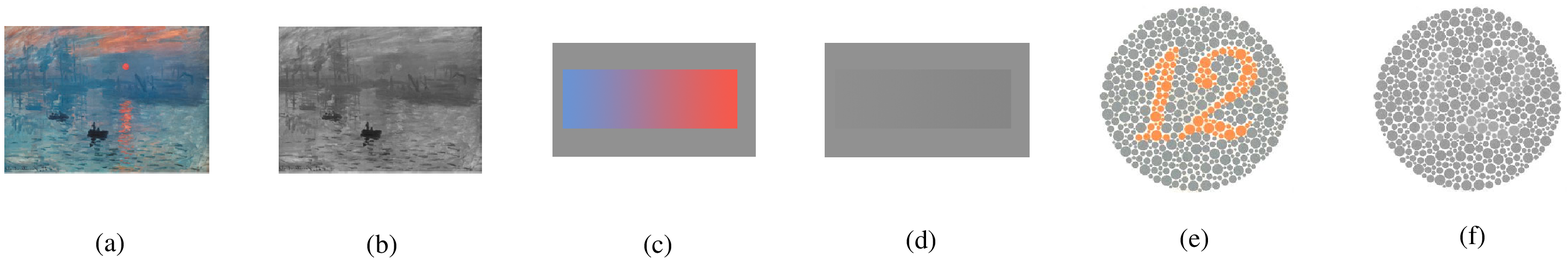}
\caption{Illustration of conventional color-to-grayscale image conversion.
    (a), (c) and (e) : Original Image. (b), (d) and (f) : Grayscale version obtained using standard NTSC rule}.
\label{issue}
\end{figure*}

For example, the pattern represented by the variation of color in the input color image (Fig.\ref{issue} (c)) is missed, when converted to gray image (Fig.\ref{issue} (d)) by the standard and most commonly used NTSC rule which combines the red, blue and green channel with the ratio 0.3: 0.6: 0.1 respectively \cite{Y.C.Faroudja1988}. Similarly, there is absence of sun and number information in the converted gray scale images (Fig.\ref{issue} (b) \& (f)). Therefore, many methods were proposed to convert the color image to grayscale with the preservation of luminance and color contrast information \cite{Liu2013,Rasche2005,Kim2009,Amy.A.Gooch2005,CeWuLu2012}. The description and the findings of the benchmark image decolorization techniques exist in the literature is given in Table \ref{exist}.

\begin{sidewaystable}[htbp]
\caption{Description of existing methods}
\label{exist}
\begin{center}
\scalebox{0.85}
{
\begin{tabular}{|c|c|l|l|}
\hline
Authors                                                                                & Dataset Used                                                                                     & \multicolumn{1}{c|}{Method Description}                                                                                                                               & \multicolumn{1}{c|}{Findings}                                                                                                                  \\ \hline
CIE \cite{M.Cadik2008}                                                               & \begin{tabular}[c]{@{}c@{}}Color-to-gray \\ benchmark dataset\end{tabular}                       & \begin{tabular}[l]{@{}l@{}} The luminance information ( Y ) of the CIE XYZ space is used \\to achieve the grayscale version of the original color image\end{tabular}                                           & \begin{tabular}[l]{@{}l@{}}This method fails in the case of images with isoluminous colors\end{tabular}                                     \\ \hline
\begin{tabular}[c]{@{}c@{}}Grundland \\and Dodgson \cite{Grundland2007}\end{tabular} & Color photographs                                                                                & \begin{tabular}[l]{@{}l@{}} The luminance information is added to the constant amount of \\chrominance information to enhance the grayscale contrast                                                \end{tabular} & \begin{tabular}[l]{@{}l@{}} The performance of this method is limited to the \\color discrimination present in the original image \end{tabular}                                            \\ \hline
\begin{tabular}[c]{@{}c@{}}Rasche et al. \cite{Rasche2005}\end{tabular}            & \begin{tabular}[c]{@{}c@{}}Color-to-gray \\ benchmark dataset \\ and natural images\end{tabular} & \begin{tabular}[l]{@{}l@{}} A constrained multivariate optimization technique is used\\ to preserve the color contrast of the original image    \end{tabular}              &\begin{tabular}[l]{@{}l@{}}                                      Computational costs of the optimization technique results in \\artifacts in the natural images with continuous tones \end{tabular}                              \\ \hline
\begin{tabular}[c]{@{}c@{}}Smith et al. \cite{Smith2008}\end{tabular}              & \begin{tabular}[c]{@{}c@{}}Color-to-gray \\ benchmark dataset \\ and natural images\end{tabular} &\begin{tabular}[l]{@{}l@{}} A global mapping based on Helmholtz-Kohlrausch effect \\along with the local chromatic contrast enhancement is \\employed in this color-to-grayscale conversion technique\end{tabular}   & \begin{tabular}[l]{@{}l@{}}This technique relies on user control and also,\\ the visual appearance of constant color region is affected  \end{tabular}                                      \\ \hline
\begin{tabular}[c]{@{}c@{}}Bala \cite{M.Cadik2008}\end{tabular}                    & \begin{tabular}[c]{@{}c@{}}Color-to-gray \\ benchmark dataset\end{tabular}                       &\begin{tabular}[l]{@{}l@{}} High-frequency chrominance information is embedded \\into the luminance to preserve the chrominance edges locally   \end{tabular}                                                       & \begin{tabular}[l]{@{}l@{}}The method is not applicable to complex images  \end{tabular}                                                                                                  \\ \hline
\begin{tabular}[c]{@{}c@{}}Neumann et al. \cite{Neumann2007}\end{tabular}          & \begin{tabular}[c]{@{}c@{}}Color-to-gray \\ benchmark dataset\end{tabular}                       & \begin{tabular}[l]{@{}l@{}}Consistent color gradients are selected locally and 2D \\fast integration is performed to attain the grayscale image    \end{tabular}                       & \begin{tabular}[l]{@{}l@{}}The details of the image may be lost by the unpredictable\\ behaviour in inconsistent regions of the gradient field     \end{tabular}                            \\ \hline
\begin{tabular}[c]{@{}c@{}}Gooch et al. \cite{Amy.A.Gooch2005}\end{tabular}          & \begin{tabular}[c]{@{}c@{}}Color-to-gray\\  benchmark dataset\end{tabular}                       & \begin{tabular}[l]{@{}l@{}} An objective function based on the local contrast between \\ pixel pairs is minimized using optimization technique\\ to preserve the color contrast of the original image  \end{tabular}   & \begin{tabular}[l]{@{}l@{}}Computationally expensive and the quality of the output image  \\ depends on the user input parameters\end{tabular} \\ \hline
\begin{tabular}[c]{@{}c@{}}Lu et al. \cite{CeWuLu2012}\end{tabular}                    & \begin{tabular}[c]{@{}c@{}}Color-to-gray \\ benchmark dataset \\ and natural images\end{tabular}                                             & \begin{tabular}[l]{@{}l@{}} Bimodal energy function is used to attain a flexible contrast\\ preserving constraint   \end{tabular}                                                                                 & Complex optimization steps increases the cost of computation                                                                                   \\ \hline
\begin{tabular}[c]{@{}c@{}}Lu et al. \cite{Lu2012}\end{tabular}                   & \begin{tabular}[c]{@{}c@{}}Color-to-gray \\ benchmark dataset \\ and natural images\end{tabular}                                              & \begin{tabular}[l]{@{}l@{}} An effective and very fast optimization based approach\\is used to maximally preserve the color contrast  \end{tabular}                                                                      & This method fails in the case of color blindness test images
\\ \hline
\begin{tabular}[c]{@{}c@{}}Sowmya et al. \cite{Sowmya2017}\end{tabular}                   & \begin{tabular}[c]{@{}c@{}}Color-to-gray \\ benchmark dataset \\ and natural images\end{tabular}                                              & \begin{tabular}[l]{@{}l@{}} An effective singular value decomposition (SVD) based \\ approach is used to incorporate the significant chrominance\\ information in the converted grayscale images.  \end{tabular}                                                                      & \begin{tabular}[l]{@{}l@{}} This method fails to preserve the contrast due to the fixed value\\ of the weightage parameter \end{tabular}                                                                                                                                             \\ \hline
\end{tabular}}
\end{center}
\end{sidewaystable}

Most of the existing methods for color-to-grayscale image conversion is based on optimization techniques which may be computationally complex \cite{Liu2013,Rasche2005,Kim2009,Amy.A.Gooch2005,CeWuLu2012}. Therefore, simple image decolorization technique using singular value decomposition (SVD) was proposed \cite{Sowmya2017}. Recently, an algorithm called color-to-gray structure similarity (C2G-SSIM) is developed to measure the level of chrominance and structure information incorporated in the converted grayscale images \cite{Ma2015}. In \cite{Ma2015}, one of the applications of C2G-SSIM is demonstrated as the parameter tuning of color-to-grayscale image conversion algorithms motivated to use the C2G-SSIM along with the SVD to develop an effective and perceptually improved method for image decolorization, which preserves the contrast information. SVD algorithm is used in our proposed work to reconstruct the chrominance planes by using the significant chrominance information captured through the singular values and the corresponding Eigenvectors.

Unlike the existing SVD based image decolorization method \cite{Sowmya2017}, the proposed framework avoids the constant weight of the chrominance information added to the luminance. Instead, the weightage of the chrominance information added to the luminance is computed using the C2G-SSIM. In \cite{Mandar}, the performance of the scene classification is improved by modeling the dense scale invariant features using gaussian mixtures and adapted gaussian mixture models. Later, the deep features to improve the scene recognition task are learnt by convolutional neural networks (CNN) using color images \cite{Zhou2014}. The performance of the scene classification system is improved by the combination of deep belief networks and support vector machines trained using the grayscale images with better chrominance representations \cite{SowmyaDBN}. Therefore, the objective of the present work is to verify the effectiveness of the chrominance information incorporated by the proposed image decolorization technique by applying the developed color-to-grayscale image conversion algorithm in the experiments for scene classification. Also, the performance of the scene classification system must be improved by the combination of the proposed image decolorization technique with the standard color-to-grayscale image conversion method.

The proposed novelties of the present work are as follows:

\begin{itemize}
\item The new method for effective and perceptually improved image decolorization method based on SVD and C2G-SSIM is proposed.
\item The role of significant and relevant chrominance information with the preservation of contrast details incorporated by the proposed method is applied for scene classification.
\item A methodology is adopted for overall improvement in the scene classification performance by combining the models developed using the perceptually better grayscale images obtained from the proposed image decolorization technique with the most commonly used image decolorization method.
\end{itemize}

The organization of the paper is as follows: The basis for the present work is discussed in section 2. Section 3 presents the proposed method followed by experimental results and analysis in section 4. Section 5 contains the summary and conclusion of the present work.

\section{Basis for the present work}
The basis for the present work is discussed in this section.

The C2G-SSIM index is computed by the following three stages, namely (i) color space transformation, (ii) measurement of luminance (L), contrast (C) and structure (S) information \& (iii) combination of the measured contents based on the type of input color image \cite{Ma2015}. In the first stage, the input color image and the converted grayscale image are transformed to CIEL*a*b* color space. The second stage involves the computation of similarity measurement of luminance, contrast and structure information between the reference color image and the converted gray image in CIEL*a*b* space. The luminance plane (of the reference color image and the converted gray image) and the chrominance planes of the reference color image is processed as windows of uniform patch size with two-dimensional Gaussian filter centered at a pixel location for every window. The luminance measure $L({x_c})$ is computed using Eq. (\ref{luminance}) \cite{Ma2015}.

\begin{equation}
L({x_c})\, = \,\frac{{2\,{u_f}\,({x_c})\,{u_g}\,({x_c})\, + \,{C_1}\,\,}}{{{u_f}\,({x_c}){\,^2}\, + \,{u_g}\,{{({x_c})}^2}\, + \,{C_1}}}\
\label{luminance}
\end{equation}

where ${u_f}\,({x_c})$ is the mean luminance computed for the luminance plane of the reference color image in CIEL*a*b* color space at the pixel location $x_c$. ${u_g}\,({x_c})$ is the mean luminance computed for the luminance plane of the converted gray image in CIEL*a*b* color space at the same pixel location $x_c$. C1 is a small positive stabilizing constant.

The chrominance measure $C({x_c})$ is computed using the Eq. (\ref{chrominance}) \cite{Ma2015}.

\begin{equation}
C({x_c})\, = \,\frac{{2\,{d_f}\,({x_c})\,{d_g}\,({x_c})\, + \,{C_2}\,\,}}{{{d_f}\,({x_c}){\,^2}\, + \,{d_g}\,{{({x_c})}^2}\, + \,{C_2}}}\
\label{chrominance}
\end{equation}

where ${d_f}\,({x_c})$ is the weighted mean color difference from its surroundings computed by considering the luminance and the chrominance details of the reference color image in CIEL*a*b* color space at the pixel location $x_c$. ${d_g}\,({x_c})$ is the mean gray tone difference computed at the same pixel location $x_c$ for the luminance plane of the converted gray image in CIEL*a*b* color space. C2 is a small positive stabilizing constant.

The structure measure $S({x_c})$ is computed using the Eq. (\ref{structure}) \cite{Ma2015}.

\begin{equation}
S({x_c})\, = \,\frac{{{\sigma _{fg}}\,({x_c})\,\, + \,{C_3}\,\,}}{{{\sigma _f}\,({x_c})\,{\sigma _g}\,({x_c})\, + \,{C_3}}}\
\label{structure}
\end{equation}

where ${\sigma _f}\,({x_c})$ is the standard deviation of the weighted mean color difference from its surroundings computed by considering the luminance and the chrominance details of the reference color image in CIEL*a*b* color space at the pixel location $x_c$. ${\sigma _g}\,({x_c})$ is the standard deviation of the mean gray tone difference computed at the same pixel location $x_c$ for the luminance plane of the converted gray image in CIEL*a*b* color space. ${\sigma _{fg}}\,({x_c})$ is the cross correlation computed between ${\sigma _f}\,({x_c})$ and ${\sigma _g}\,({x_c})$ at the same pixel location $x_c$. C3 is a small positive stabilizing constant.

Thus, the overall C2G-SSIM index $q({x_c})$ is computed by combining all the three contents, namely luminance $L({x_c})$, contrast $C({x_c})$ and structure $S({x_c})$ at the pixel location $x_c$ using the Eq. (\ref{C2G-SSIM}) \cite{Ma2015}.

\begin{equation}
q({x_c})\, = \,L{({x_c})^\alpha }\, \times C{({x_c})^\beta }\, \times S{({x_c})^\gamma }\
\label{C2G-SSIM}
\end{equation}

where $\alpha\,>\,0$, $\beta\,>\,0$, $\gamma\,>\,0$. In \cite{Ma2015}, the expression for C2G-SSIM index is simplified by setting $\beta = \gamma = 1$. And, $\alpha = 1$ for photographic images and $\alpha = 0$ for synthetic images.

\section{Proposed Method}

\begin{figure}[htbp]
\centering
\includegraphics[width=6in,height=2in]{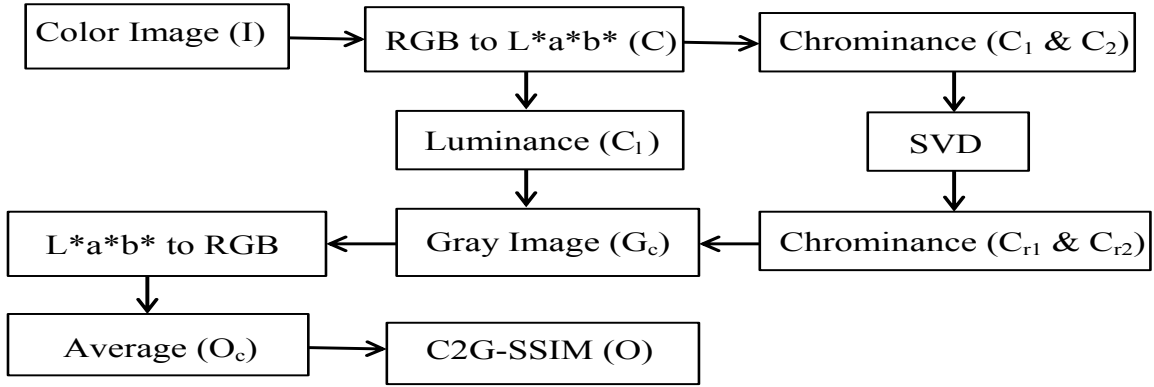}
\vspace{0.5cm}
\caption{Block diagram of the proposed methodology}.
\label{block}
\end{figure}

An image decolorization method is proposed to perceptually improve the performance of the color-to-grayscale image conversion using C2G-SSIM and SVD. The algorithm of the proposed image decolorization is discussed below. The block diagram of the proposed methodology shown in Fig.\ref{block} is explained below.

\begin{enumerate}
\item Input Color Image: $I\in{R^{h\, \times \,w\, \times \,3}}$, where $h$ and $w$ refers to the row and column dimension of the input color image. The third dimension refers to the three color planes namely red, green and blue.
\item Image Transformation: $C = T\,(I)$ , where $C$ represents the transformed image in CIEL*a*b* color space which is the better suited color space for image editing \cite{Sowmya2017}. $T$ is the color space transformation function which converts the input color image in RGB space to CIEL*a*b* space.
\item Singular Value Decomposition: Each of the chrominance planes of the transformed image is subjected to singular value decomposition. $[{U_i}\, {S_i}\, {V_i}^{'}] = svd ({C_{i\,}})\forall \,i = 1,2.$, where $C_i$ refers to the first and the second chrominance planes of the transformed color image for $i=1 \,\&\, 2$ respectively.$U_i$, and $V_i$ contains the Eigen vectors corresponding to the input chrominance planes. $S_i$ contains the Eigen values corresponding to the input chrominance planes.
\item Chrominance Planes Reconstruction ($Cr$): Each of the chrominance planes is reconstructed using the Eigen values and the corresponding Eigen vectors selected based on the rank of the input chrominance plane. $C{r_i}\,\, = \,\,U{r_i}\, \times \,\,S{r_i}\, \times \,\,V{r_i}^{'}\,\,,\,\,\forall \,i\, = 1,\,2.$.
\item Gray/Decolorized Image($G$): The gray image is obtained using the equation (\ref{gray})

\begin{equation}
{G_c}(k)\, = \,{C_l}\, + \,c\, \times \,\sum\limits_{i = 1}^2 {{C_{ri}}} (k)\,\,\,,\,\,\forall \,k\, = \,1\,\,to\,\,h \times w
\label{gray}
\end{equation}

\item Transformation to RGB color space \& average computation: $O_c = T{'}\,(G_c)$, where $T{'}$ transforms the gray image to RGB space with only the luminance information. The average of three planes in RGB space ($O_cR$, $O_cG$, $O_cB$ ) is computed using the equation (\ref{avg}) to form the final gray or decolorized image \cite{Sowmya2017}.

\begin{equation}
{O_c}^{'} = \,({O_c}R\, + \,{O_c}G\, + {O_c}B)\,/3
\label{avg}
\end{equation}

Though the output gray image obtained by the existing SVD image decolorization technique is comparable with all the benchmark algorithms, the weightage parameter of the chrominance information added to the luminance is fixed experimentally as 0.25 \cite{Sowmya2017}. The fixed value of weightage parameter fails to preserve the contrast in the converted grayscale image which affects the perceptual quality of the converted image.

\begin{figure*}
\centering
\includegraphics[width=\textwidth,height=2.25in]{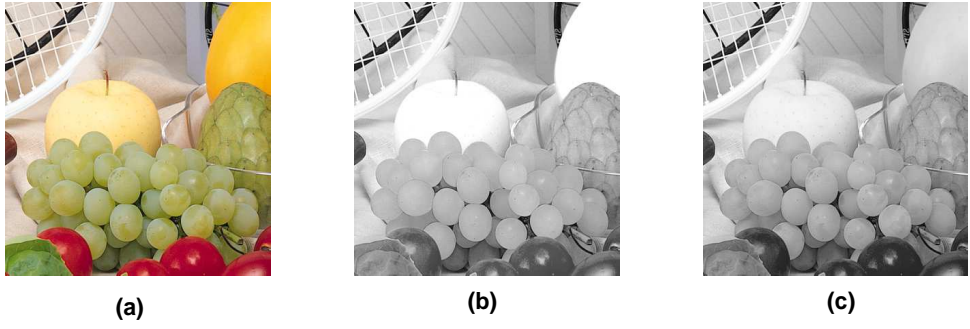}
\caption{Illustration of color-to-grayscale image conversion by SVD image decolorization method \cite{Sowmya2017}.
    (a) : Original Image. Grayscale version obtained using (b) : SVD image decolorization (c) : Proposed Method}.
\label{SVDissue}
\end{figure*}

For example, the color contrast information present in the original color image (region where apple and orange in present) is not preserved in the grayscale image converted by the existing SVD based image decolorization (Fig.\ref{SVDissue}(b)). Fig.\ref{SVDissue} (c) shows perceptually improved  image decolorization from Fig.\ref{SVDissue} (a) as compared  to the  grayscale image in Fig.\ref{SVDissue} (b) which is obtained using the conventional  SVD  based  method.

The constant value $c$ is chosen using the measure designed to measure the significant components of color-to-grayscale image conversion such as luminance, structure and contrast information known as C2G-SSIM. The mathematical representation corresponding to our proposed method for color-to-grayscale image conversion is given by:

\begin{equation}
O\, = \,\mathop {\arg \max }\limits_c \,(\,q\,(I,\,{O_c{'}}))\
\label{C2G-SSIM}
\end{equation}

where $q$ is the C2G-SSIM index computed using the Eq. (\ref{C2G-SSIM}).

\end{enumerate}

\section{Experimental Results and Analysis}
The proposed image decolorization technique to improve the perceptual quality of the image decolorization is evaluated as follows:

\begin{enumerate}
\item Objective evaluation using quality assessment metric for image decolorization namely color-to-gray structure similarity index (C2G-SSIM) \cite{Ma2015}.
\item Role of incorporating significant chrominance information by the proposed perceptually improved image decolorization method for scene classification.
\end{enumerate}

\subsection{Objective evaluation of the proposed perceptually improved SVD image decolorization using C2G-SSIM metric}

The proposed method to improve the perceptual quality of the image decolorization technique using C2G-SSIM is experimented on two different benchmark datasets used for color-to-grayscale image conversion. The first dataset known as Cadik, contains 25 different color images which includes isoluminous (images with uniform light intensity values) color images \cite{M.Cadik2008}. The second dataset is COLOR250, which contains 250 different color images \cite{Lu2014}. COLOR250 dataset contains generic color images along with the isoluminous pie chart color images. The proposed technique is experimented for 20 different values of the weight parameter ($c$) ranging from 0.05 to 1.0 with the incremental step value of 0.05.

The C2G-SSIM index is computed for the different values of $c$ for all the images in Cadik and COLOR250 datasets for which the samples are tabulated in Table \ref{c-values}. The experimental analysis computed for all the images in both the dataset shows that the best C2G-SSIM index is obtained for different value of $c$ other than $c=0.25$ as fixed in SVD decolorization algorithm for color-to-grayscale image conversion. For instance, the color images from Cadik dataset namely `balls0\_color' and `tulips' have 0.9056 and 0.7860 respectively as C2G-SSIM index for $c=0.05$, which is better than the index obtained as 0.8553 and 0.7057 repsectively for $c=0.25$, as fixed in SVD decolorization method. Similar analysis are observed for all the images in both the datasets. For example, color image namely `im\_8' and `im\_10' in COLOR250 dataset have C2G-SSIM index as 0.8964 and 0.8090 for $c=0.15$, which is better than the index obtained for the same images at $c=0.25$. (SVD decolorization).

\begin{table*}[htbp]
\centering
\caption{C2G-SSIM index computed for different values of parameter (c) for the images in Cadik and COLOR250 dataset \cite{M.Cadik2008}, \cite{Lu2014}}
\label{c-values}
\vspace{0.25cm}
\scalebox{0.68}{
\begin{tabular}{|l|l|l|l|l|l|l|l|l|l|l|l|l|l|l|}
\hline
\multirow{2}{*}{Image\_name} & \multicolumn{14}{c|}{c-value}                                                                                               \\ \cline{2-15}
                             & 0.05            & 0.15   & 0.25   & 0.35   & 0.45   & 0.55   & 0.65            & 0.7 &0.75             & 0.8 &0.85    & 0.9 &0.95    & 1.0    \\ \hline
balls0\_color                & \textbf{0.9056} & 0.8866 & 0.8553 & 0.8153 & 0.7664 & 0.7110 & 0.6557          & 0.6285 &0.6014          & 0.5744 &0.5478 & 0.5217 &0.4963 & 0.4718 \\ \hline
ColorsPastel                 & 0.7314          & 0.7374 & 0.7519 & 0.7769 & 0.7858 & 0.8023 & \textbf{0.8170} & 0.8168 &8143         & 0.8112 & 8076 & 0.8042 &0.8008 & 0.7967 \\ \hline
IM2-color                    & 0.8094          & 0.8474 & 0.8883 & 0.9141 & 0.9305 & 0.9380 & 0.9396          & \textbf{0.9400} & 0.9398 & 0.9399 &0.9395 & 0.9381 &0.9365 & 0.9343 \\ \hline
tulips                       & \textbf{0.7860}  & 0.7538 & 0.7057 & 0.6534 & 0.6144 & 0.5967 & 0.5709          & 0.5575 &0.5444          & 0.5317 &0.5194 & 0.5072 &0.4954 & 0.4838 \\ \hline
im\_8                      & 0.8838  & \textbf{0.8964} & 0.8692 & 0.8256 & 0.7880 & 0.7526 & 0.7285          & 0.7191 &0.7115          & 0.7042 &0.6976 & 0.6914 &0.6852 & 0.6796 \\ \hline
im\_13                     & 0.9103  & 0.9154 & \textbf{0.9171} & 0.9150 & 0.9079 & 0.8883 & 0.8641          & 0.8539 &0.8445          & 0.8351 &0.8254 & 0.8156 &0.8057 & 0.7954 \\ \hline
im\_10                     & 0.8048  & \textbf{0.8090} & 0.8075 & 0.8011 & 0.7873 & 0.7633 & 0.7245          & 0.7054 &0.6879          & 0.6717 &0.6563 & 0.6417 &0.6279 & 0.6143 \\ \hline

\end{tabular}}
\end{table*}

\begin{figure*}
\centering
\includegraphics[width=\textwidth,height=3in]{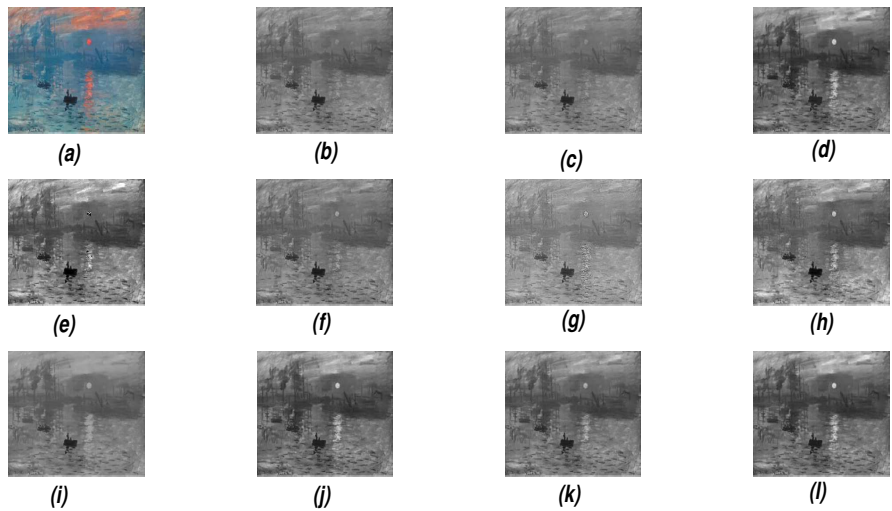}
\caption{Visual Comparison of the proposed color-to-grayscale image conversion against the existing benchmark techniques for a sample image (Sunrise312) in Cadik dataset \cite{M.Cadik2008} (a): Input Color Image. (b)-(k): Grayscale version obtained using the existing benchmark methods namely rgb2gray, CIE\_Y, Decolor, Rasche, Smith, Bala, Neuman, Gooch, RT-CP, SVD respectively. (l): Gray image obtained using the proposed technique}.
\label{sunrise}
\end{figure*}

\begin{figure*}
\centering
\includegraphics[width=\textwidth,height=3in]{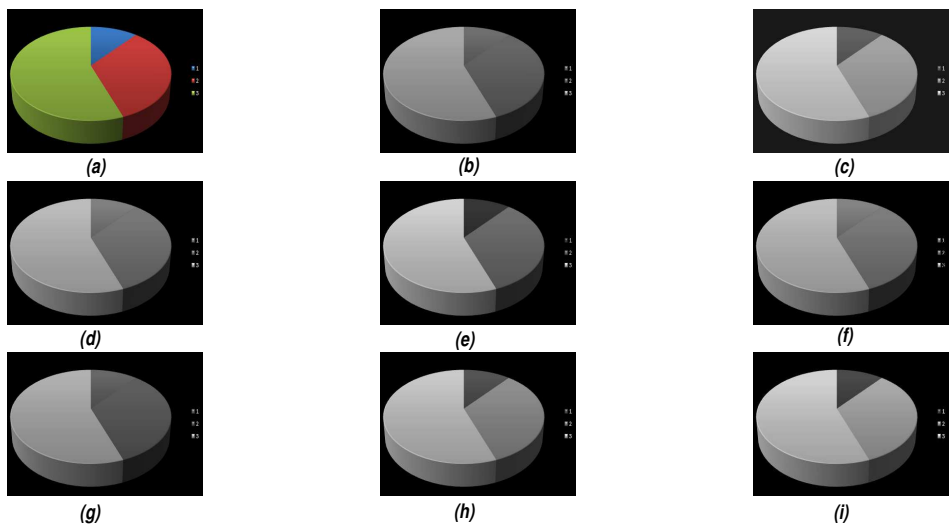}
\caption{Visual Comparison of the proposed color-to-grayscale image conversion against the existing benchmark techniques for a sample image (im\_203) in COLOR250 dataset \cite{Lu2014} (a): Input Color Image. (b)-(h): Grayscale version obtained using the existing benchmark methods namely rgb2gray, RT-CP, CIE\_Y, Decolor, Smith, Gooch, SVD respectively. (i): Gray image obtained using the proposed technique}.
\label{im203}
\end{figure*}

The visual comparison of the gray image `Sunrise312' obtained using the proposed technique and all the benchmark technique is shown in Fig.\ref{sunrise}. The gray image obtained using the proposed technique (Fig.\ref{sunrise} (l)) has better contrast along with the structure content as in the original color image (Fig. \ref{sunrise} (a)). Hence, the C2G-SSIM index computed for the gray image converted using the proposed technique is better than all other existing benchmark color-to-gray image conversion techniques. The similar observation is also recorded for the pie chart image namely `im\_203' present in COLOR250 dataset. The variation in the colors present in the input color image (Fig.\ref{im203} (a)) is missed in the corresponding gray images obtained using the existing color-to-gray conversion algorithms (Fig.\ref{im203} (b), (d), (f) \& (g)). Whereas, in the case of gray image obtained using the proposed technique (Fig.\ref{sunrise} (i)), the color variation present in the input color image is represented with better contrast and structure in gray intensity level. Hence, the C2G-SSIM index computed for the gray image (im\_203) obtained using the proposed technique is 0.9927 (Table \ref{C2G-SSIM-compare1}), which is better than the index computed for all the existing benchmark algorithms. The reported trend of the proposed color-to-grayscale image conversion method holds good for all the sample images taken from COLOR250 dataset as tabulated in Table \ref{C2G-SSIM-compare1}.

The proposed color-to-grayscale image conversion based on the variation of the parameter ($c$) using C2G-SSIM index is compared against the existing benchmark techniques for all the images available in Cadik and COLOR250 datasets. The C2G-SSIM index computed for the sample images from both the datasets for the proposed method and existing benchmark techniques for color-to-grayscale image conversion are tabulated in Table \ref{C2G-SSIM-compare} \& \ref{C2G-SSIM-compare1}. The C2G-SSIM value computed for the gray image obtained using the proposed decolorization algorithm is better than all the existing benchmark techniques. For example, in case of `Sunrise312' (Cadik dataset) in Table \ref{C2G-SSIM-compare}, the C2G-SSIM index obtained for the gray image converted using the proposed technique is 0.8898 which is better than the values obtained for the grayscale image converted using all the benchmark algorithms.

\begin{table*}[htbp]
\centering
\caption{Comparison of the proposed method for color-to-gray image conversion against the existing benchmark techniques for Cadik dataset based on C2G-SSIM quality index}
\label{C2G-SSIM-compare}
\vspace{0.25cm}
\scalebox{0.65}{
\begin{tabular}{|l|c|c|c|c|c|c|c|c|c|c|c|c|}
\hline
Image\_name  & \multicolumn{1}{l|}{rgb2gray \cite{Y.C.Faroudja1988}} & \multicolumn{1}{l|}{CIE\_Y \cite{M.Cadik2008}} & \multicolumn{1}{l|}{Decolor \cite{M.Cadik2008}} & \multicolumn{1}{l|}{Rasche \cite{M.Cadik2008}} & \multicolumn{1}{l|}{Smith \cite{M.Cadik2008}} & \multicolumn{1}{l|}{Bala \cite{M.Cadik2008}} & \multicolumn{1}{l|}{Neuman \cite{M.Cadik2008}} & \multicolumn{1}{l|}{Gooch \cite{M.Cadik2008}} & \multicolumn{1}{l|}{CP \cite{M.Cadik2008}} & \multicolumn{1}{l|}{RT-CP \cite{Lu2012}} & \multicolumn{1}{l|}{SVD \cite{Sowmya2017}} & \multicolumn{1}{l|}{Proposed} \\ \hline
155\_5572    & 0.8554                                                  & 0.8589                                           & 0.8424                                            & 0.6629                                           & 0.8584                                          & 0.8233                                         & 0.8212                                           & 0.8277                                          & 0.7749                                     & 0.8303                                     & 0.8147                                       & \textbf{0.8647}               \\ \hline
34445        & 0.8638                                                  & 0.8595                                           & 0.8356                                            & 0.8556                                           & 0.8551                                          & 0.7681                                         & 0.7279                                           & 0.8592                                          & 0.8246                                     & 0.8540                                     & 0.8579                                       & \textbf{0.8659}               \\ \hline
portrait\_4v & 0.9289                                                  & 0.9225                                           & 0.9269                                            & 0.9228                                           & 0.9270                                          & 0.8849                                         & 0.8901                                           & 0.9294                                          & 0.8836                                     & 0.9318                                     & 0.9234                                       & \textbf{0.9323}               \\ \hline
Sunrise312   & 0.8576                                                  & 0.8455                                           & 0.8721                                            & 0.8247                                           & 0.8567                                          & 0.8337                                         & 0.8312                                           & 0.8688                                          & 0.7473                                     & 0.8810                                     & 0.8839                                       & \textbf{0.8898}               \\ \hline
watch        & 0.9611                                                  & 0.9506                                           & 0.9519                                            & 0.9569                                           & 0.9587                                          & 0.8904                                         & 0.8112                                           & 0.9612                                          & 0.9277                                     & 0.9628                                     & 0.9611                                       & \textbf{0.9642}               \\ \hline
%success rate& 2                                                  & 1                                          & 0                                          & 1                                          & \textbf{7}                                         & 0                                         & 0                                          & 0                                         & \textbf{7}                                     & 2                                     & 1                                       & 4               \\ \hline
\end{tabular}
}
\end{table*}

\begin{table*}[htbp]
\centering
\caption{Comparison of the proposed method for color-to-gray image conversion against the existing benchmark techniques for COLOR250 dataset based on C2G-SSIM quality index}
\label{C2G-SSIM-compare1}
\vspace{0.25cm}
\scalebox{0.85}{
\begin{tabular}{|l|c|c|c|c|c|c|c|c|c|}
\hline
Image\_name  & \multicolumn{1}{l|}{rgb2gray \cite{Y.C.Faroudja1988}} & \multicolumn{1}{l|}{RT-CP \cite{Lu2012}} & \multicolumn{1}{l|}{CP \cite{M.Cadik2008}} & \multicolumn{1}{l|}{CIE\_Y \cite{M.Cadik2008}} & \multicolumn{1}{l|}{Decolor \cite{M.Cadik2008}} & \multicolumn{1}{l|}{Smith \cite{M.Cadik2008}}
& \multicolumn{1}{l|}{Gooch \cite{M.Cadik2008}}  & \multicolumn{1}{l|}{SVD \cite{Sowmya2017}} & \multicolumn{1}{l|}{Proposed} \\ \hline
im\_18 & 0.8483                                                  & 0.7815                                           & 0.8372                                            & 0.8474                                           & 0.8129                                          & 0.8469                                         & 0.8186                                           & 0.8332                                          & \textbf{0.8580}                                   \\ \hline

im\_38 & 0.9061                                                  & 0.8732                                          & 0.8802                                            & 0.9176                                           & 0.8749                                         & 0.9174                                        & 0.9111                                           & 0.8971                                          & \textbf{0.9236}                                   \\ \hline

im\_66 & 0.9610                                                  & 0.9625                                          & 0.9622                                            & 0.9642                                          & 0.9408                                         & 0.9651                                       & 0.9606                                           & 0.9634                                          & \textbf{0.9708}                                   \\ \hline

im\_89 & 0.9229                                                  & 0.9194                                         & 0.9187                                            & 0.9329                                         & 0.8746                                         & 0.9323                                       & 0.9256                                           & 0.9256                                         & \textbf{0.9334}                                   \\ \hline

im\_117 & 0.9579                                                 & 0.9497                                         & 0.9511                                            & 0.9553                                        & 0.9128                                         & 0.9544                                      & 0.9556                                           & 0.9502                                        & \textbf{0.9607}                                   \\ \hline

im\_141 & 0.9685                                                & 0.9633                                         & 0.9649                                            & 0.9569                                        & 0.9562                                         & 0.9577                                      & 0.9626                                           & 0.9503                                        & \textbf{0.9707}                                   \\ \hline

im\_160 & 0.7585                                                & 0.6263                                         & 0.6752                                            & 0.7596                                       & 0.7558                                         & 0.7451                                      & 0.7414                                           & 0.7160                                        & \textbf{0.7774}                                   \\ \hline

im\_175 & 0.8934                                                & 0.8979                                         & 0.8891                                            & 0.8902                                      & 0.8656                                        & 0.8897                                      & 0.8729                                           & 0.8747                                       & \textbf{0.9001}                                   \\ \hline

im\_187 & 0.8718                                                & 0.8585                                         & 0.8272                                            & 0.8851                                      & 0.7625                                        & 0.8742                                      & 0.8822                                           & 0.8850                                      & \textbf{0.8860}                                   \\ \hline

im\_203 & 0.9868                                               & 0.9914                                        & 0.9895                                            & 0.9885                                     & 0.9909                                        & 0.9883                                      & 0.9865                                         & 0.9920                                     & \textbf{0.9927}                                   \\ \hline

im\_227 & 0.9004                                               & 0.8601                                         & 0.8560                                            & 0.8966                                      & 0.9110                                        & 0.8945                                      & 0.8828                                         & 0.9243                                     & \textbf{0.9243}                                   \\ \hline

%success rate &12                                               & 18                                        & 37                                          & 29                                   & 10                                       & 25                                      & 4                                        & 2                                    & \textbf{113}                                  \\ \hline

\end{tabular}}
\end{table*}

\begin{figure*}
\centering
\includegraphics[width=\textwidth,height=2.5in]{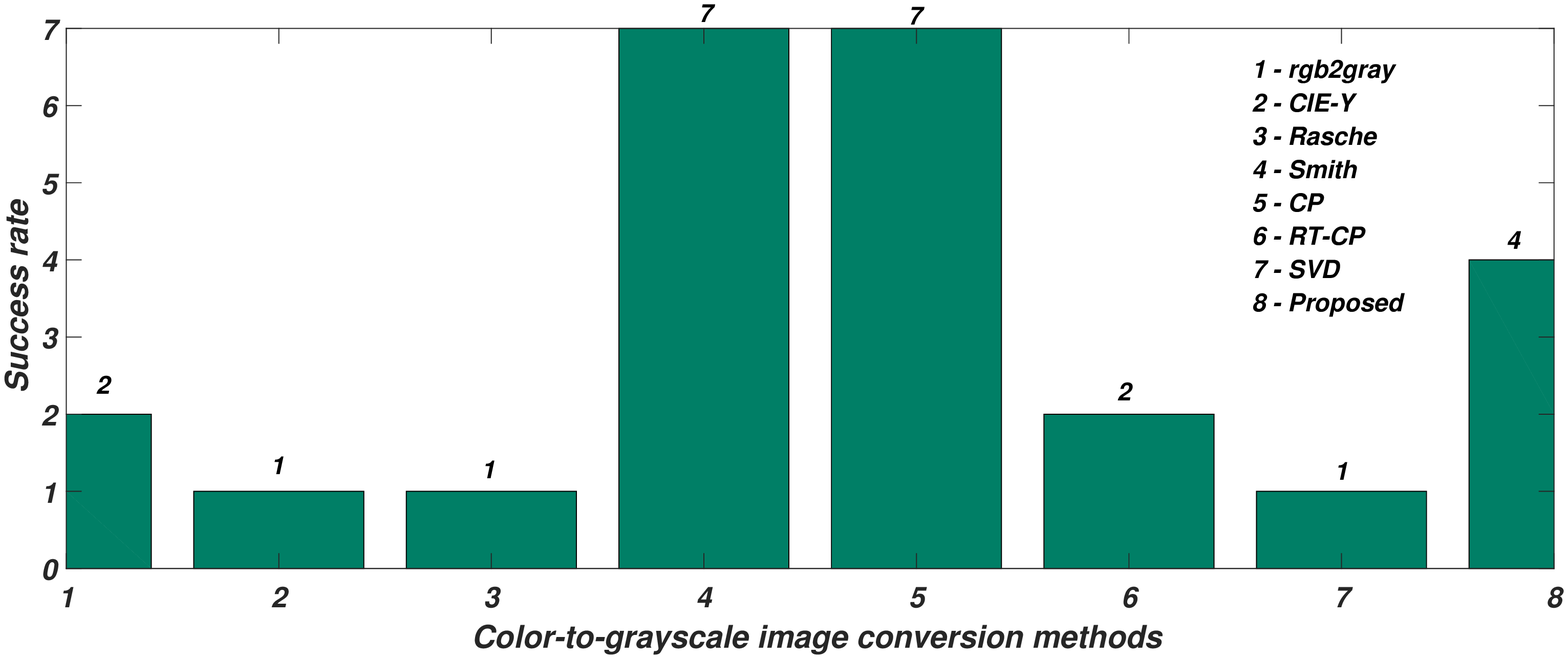}
\vspace{0.25cm}
\caption{Graphical representation of success rate of the proposed color-to-grayscale image conversion against the existing benchmark techniques for the images in Cadik dataset \cite{M.Cadik2008}}.
\label{srate25}
\end{figure*}

\begin{figure*}
\centering
\includegraphics[width=\textwidth,height=2.5in]{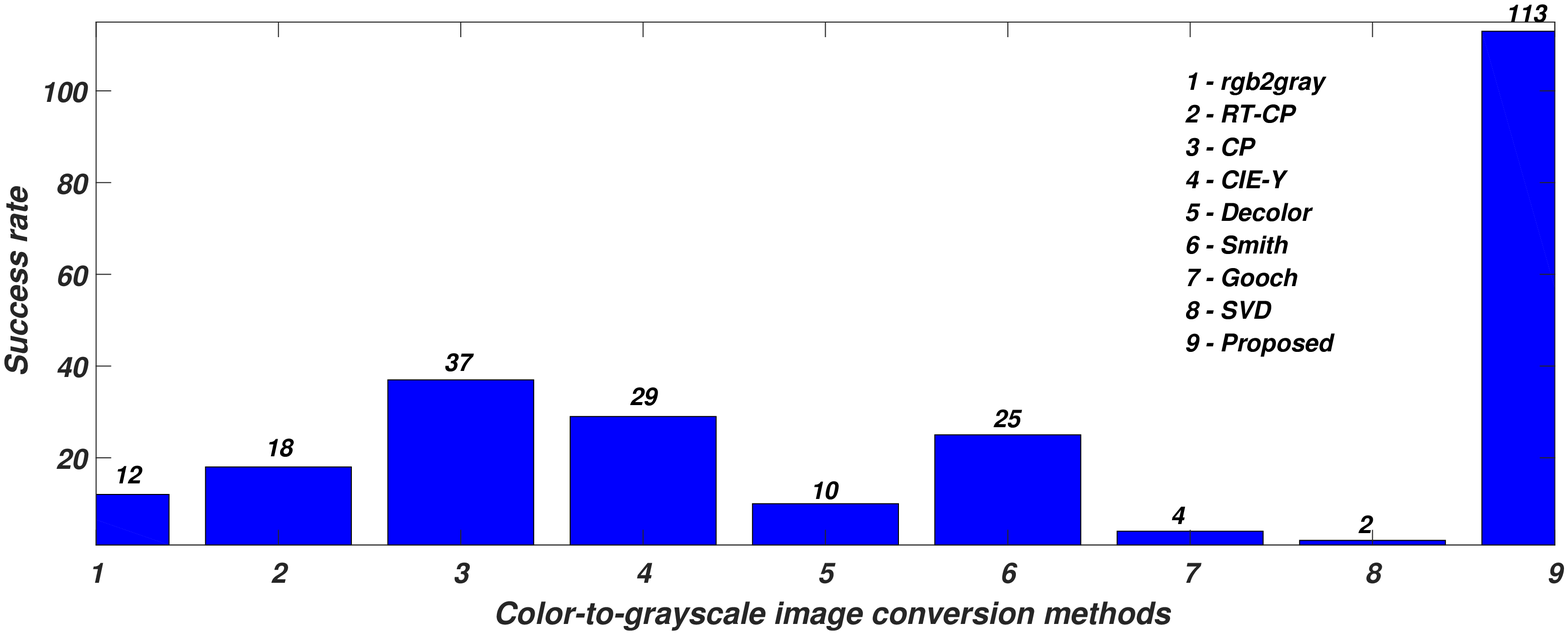}
\vspace{0.25cm}
\caption{Graphical representation of success rate of the proposed color-to-grayscale image conversion against the existing benchmark techniques for the images in COLOR250 dataset \cite{Lu2014}}.
\label{srate250}
\end{figure*}

\begin{figure*}
\centering
\includegraphics[width=\textwidth,height=3in]{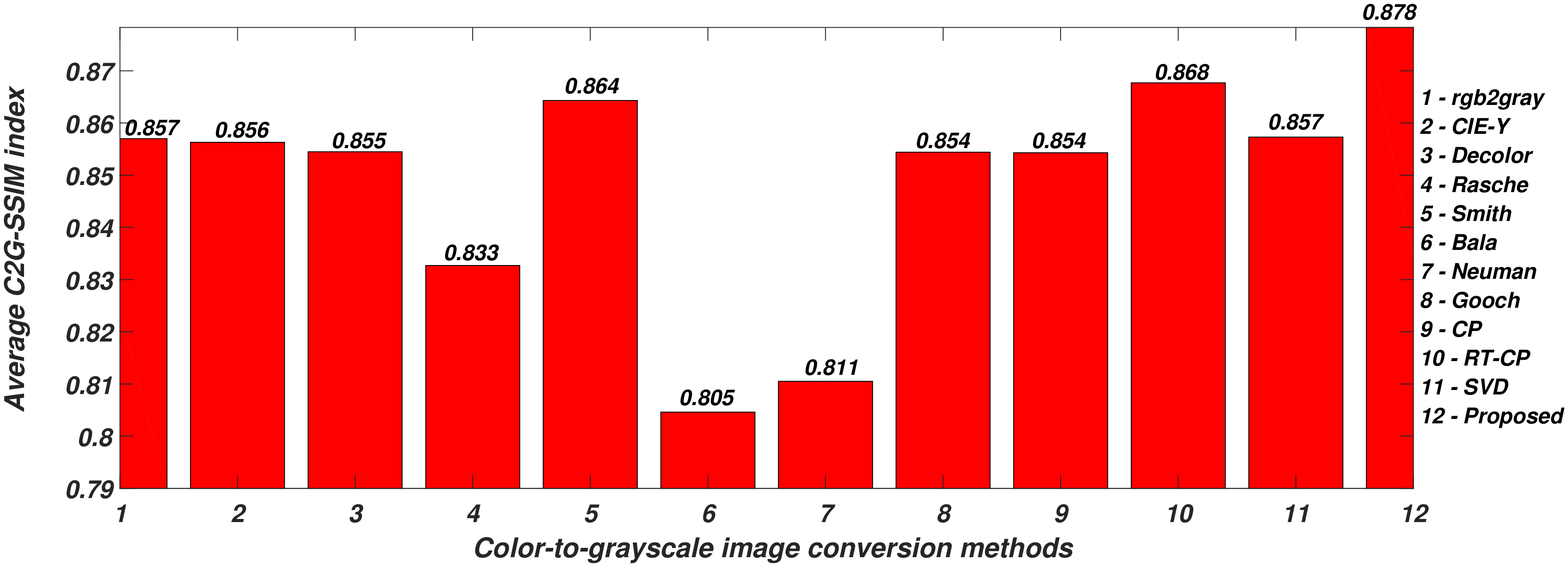}
\vspace{0.25cm}
\caption{Graphical representation of average C2G-SSIM index computed for the proposed color-to-grayscale image conversion and the existing benchmark techniques using all the images in Cadik dataset \cite{M.Cadik2008}}.

\label{average25}
\end{figure*}

\begin{figure*}
\centering
\includegraphics[width=\textwidth,height=3in]{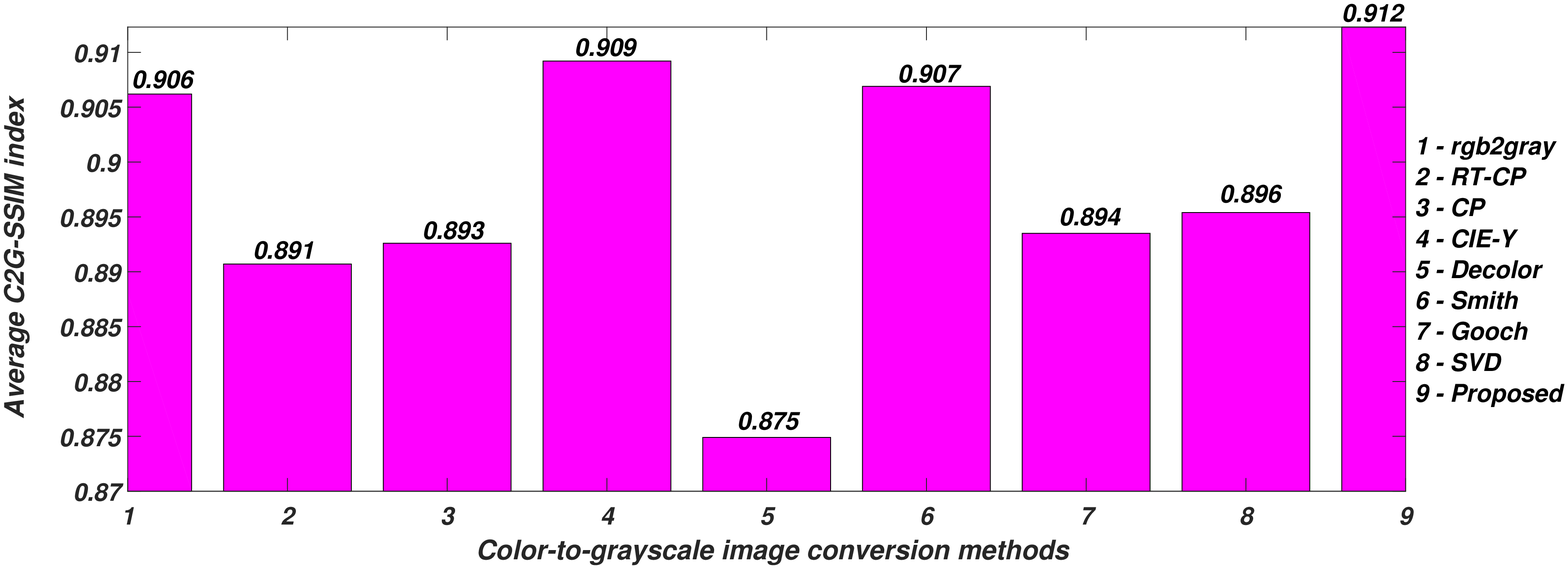}
\vspace{0.25cm}
\caption{Graphical representation of average C2G-SSIM index computed for the proposed color-to-grayscale image conversion and the existing benchmark techniques using all the images in COLOR250 dataset \cite{Lu2014}}.
\label{average250}
\end{figure*}

The success rate of a technique is defined as the number of images whose C2G-SSIM score is maximum in the corresponding technique when compared with the other existing techniques \cite{Suhre2010}. The success rate computed for the proposed technique and all the existing algorithms for color-to-grayscale image conversion is represented graphically in Fig.\ref{srate25} and Fig.\ref{srate250} for Cadik and COLOR250 datasets respectively. Out of 25 images available in Cadik dataset, none of the images obtained maximum C2G-SSIM score for the gray images obtained using the existing algorithms namely Decolor, Bala, Neuman, Gooch. Therefore, the success rate for all these techniques is 0. The success rate for the proposed technique is 4 which is better than the success rate obtained for SVD decolorization, RT-CP, Rasche, CIE\_Y and rgb2gray.

In case of COLOR250 dataset which contains 250 different color images, the success rate of the proposed technique is 113 which is greater than the success rate obtained for all other existing image decolorization algorithms. The success rate of the existing techniques namely CP and Smith is 7 in case of Cadik dataset, which is higher than that of the proposed technique (Fig.\ref{srate25}). But, in case of COLOR250 dataset, the success rate of CP and Smith are 37 and 25 respectively, which is lower than that of the proposed technique (Fig.\ref{srate250}).

The average C2G-SSIM index computed for all the images in both the dataset for the proposed technique and all the existing image decolorization algorithms are represented graphically in Fig.\ref{average25} and Fig.\ref{average250}. The average C2G-SSIM index computed for the proposed technique using all the image sin Cadik dataset is 0.878, which is better than all the existing color-to-grayscale image conversion methods. Although the success rate of CP and Smith is greater than the proposed technique in case of Cadik dataset, the average C2G-SSIM index corresponding to those two techniques are lower (0.8543 and 0.8643 respectively) than that of the proposed image decolorization technique. Similarly, in the case of COLOR250 dataset, the average C2G-SSIM index is 0.912, which is better than all the existing image decolorization techniques. The implementation of the proposed method to perceptually improve the image decolorization is available in http://nlp.amrita.edu/PerceptualC2G/index.html.

\subsection{Role of incorporating significant chrominance information by the proposed perceptually improved SVD image decolorization for scene classification}

The experimental setup to analyze the effect of significant chrominance information on scene classification is as follows: In general, the scene classification system contains three stages namely color-to-grayscale image conversion, feature extraction and classification.

\begin{enumerate}
\item Color-to-grayscale image conversion: In our experiment, the proposed color-to-grayscale image conversion to perceptually improve the existing SVD image decolorization is used in the first stage to convert the input color images to grayscale. To compare the effect of the proposed image decolorization method on scene classification, the input color images are also converted to grayscale using the most commonly used technique called the National Television Standards Committee (NTSC) rule with the ratio 0.3 : 0.6 : 0.1 of RGB respectively \cite{Y.C.Faroudja1988} and the existing method called SVD image decolorization \cite{Sowmya2017}.
\item Feature Extraction: The dense scale invariant feature transform (SIFT) which is the suitable feature space for scene classification is used in the present work \cite{Lazebnik2006, Mandar}. Each grayscale converted input image of size $256\times 256$ is divided into grids of size $16 \times 16$ with 8 pixel spacing between the grids. Hence, there are 961 grids extracted from each converted grayscale image. Each grid is represented by a SIFT feature descriptor of length 128. Thus, each grayscale converted input image is represented by feature matrix of size $961 \times 128$.
\item Classification: The two different types of classifiers namely Adapted Gaussian Mixture Models combined with Support Vector Machines (AGMM-SVM) and AGMM combined with Deep Belief Networks and SVM (AGMM-DBN-SVM) proposed by Mandar et al. \cite{Mandar} \& Sowmya et al. \cite{SowmyaDBN} respectively are used in the present experimental work.
\end{enumerate}

The experimental setup for AGMM-SVM classification system is followed from \cite{Mandar}. The number of gaussian mixtures used to adapt the mean SIFT feature matrix of each class is 1024. The mean adapted SIFT features of each class of dimension $1024 \times 128$ is used as dictionary to derive the histogram feature representation of each converted grayscale image. The adapted gaussian mixture which is closest to each of the SIFT feature descriptors of an image is computed whose dimension is $1 \times 961$ with the values ranging from 1 to number of mixtures (1024). The histogram is calculated for the above obtained feature vector. The normalized histogram vector is computed with respect to the dictionary of all the classes involved in the scene classification. The average histogram vector is computed and is fed as input to the SVM classifier.

In case of AGMM-DBN-SVM system, the average histogram feature vector corresponding to all the converted grayscale images derived from training set are used to learn the DBN network proposed in \cite{SowmyaDBN}. The DBN network used in the present experimental work consists of two hidden layers apart from the input and the output layer. The output layer of DBN are used as input features to the SVM classifier. The two hidden layers contains 30 and 99 neurons respectively for standard color-to-grayscale image conversion (NTSC rule) \cite{SowmyaDBN}. The number of neurons present in the two hidden layers for the existing SVD image decolorization are 88 and 100 respectively \cite{SowmyaDBN}. The architecture of DBN network for the proposed image decolorization is same as that of the existing SVD image decolorization method.

The standard Oliva Torralba (OT) scene dataset containing 8 different classses namely opencountry, coast, forest, highways, insidecity, mountain, street and tallbuildings is used for the experimental work \cite{OT2001, Xie2012, Lazebnik2006, Mandar, Fei-Fei2005}. 100 images from each class are used for training and all the remaining images are used for testing.

\begin{table}[htbp]
\centering
\caption{Comparison of classwise accuracy (\%) and overall accuracy (\%) for AGMM-SVM scene classification system formed using the proposed and the existing image decolorization methods}
\label{AGMM-SVM}
\vspace{0.5cm}
\scalebox{0.75}{
\begin{tabular}{|l|c|c|c|c|c|c|c|c|c|}
\hline
\multicolumn{1}{|c|}{\begin{tabular}[c]{@{}c@{}}Image Decolorization\\ Methods\end{tabular}} & \multicolumn{1}{l|}{opencountry} & \multicolumn{1}{l|}{coast} & \multicolumn{1}{l|}{forest} & \multicolumn{1}{l|}{highways} & \multicolumn{1}{l|}{insidecity} & \multicolumn{1}{l|}{mountain} & \multicolumn{1}{l|}{street} & \multicolumn{1}{l|}{tallbuilding} & \multicolumn{1}{l|}{overall accuracy} \\ \hline
NTSC rule                                                                                   & 71.9                             & 80.0                       & 94.3                        & 76.9                          & 79.3                            & 82.1                          & 90.1                        & 82.8                              & 81.8                                  \\ \hline
SVD                                                                                         & 72.3                             & 79.6                       & 94.3                        & 76.3                          & 76.9                            & 81.8                          & 89.6                        & 84.4                              & 81.6                                  \\ \hline
\begin{tabular}[c]{@{}l@{}}Proposed\\ (Modified SVD)\end{tabular}                                                                     & 71.9                             & 78.5                       & 93.9                        & 73.8                          & \textbf{80.8}                   & \textbf{83.9}                 & \textbf{90.1}               & \textbf{85.2}                     & \textbf{82.0}                         \\ \hline
\end{tabular}}
\end{table}

\begin{table}[htbp]
\centering
\caption{Comparison of classwise accuracy (\%) and overall accuracy (\%) for AGMM-DBN-SVM scene classification system formed using the proposed and the existing image decolorization methods}
\label{AGMM-DBN-SVM}
\vspace{0.5cm}
\scalebox{0.75}{
\begin{tabular}{|l|c|c|c|c|c|c|c|c|c|}
\hline
\multicolumn{1}{|c|}{\begin{tabular}[c]{@{}c@{}}Image Decolorization\\ Methods\end{tabular}} & \multicolumn{1}{l|}{opencountry} & \multicolumn{1}{l|}{coast} & \multicolumn{1}{l|}{forest} & \multicolumn{1}{l|}{highways} & \multicolumn{1}{l|}{insidecity} & \multicolumn{1}{l|}{mountain} & \multicolumn{1}{l|}{street} & \multicolumn{1}{l|}{tallbuilding} & \multicolumn{1}{l|}{overall accuracy} \\ \hline
NTSC rule                                                                                   & 90.9                             & 92.7                       & 92.1                        & 87.5                          & 93.3                           & 95.9                          & 96.4                        & 97.7                              & 93.5                                  \\ \hline
SVD                                                                                         & 91.6                             & 93.5                       & 94.7                        & 87.5                          & 91.4                            & 95.3                          & 96.9                        & 96.9                              & 93.6                                  \\ \hline
\begin{tabular}[c]{@{}l@{}}Proposed\\ (Modified SVD)\end{tabular}                                                                     & \textbf{94.2}                            & \textbf{96.9}                       & \textbf{95.2}                        & \textbf{91.3}                          & 92.8                   & \textbf{96.4}                & \textbf{97.4}               & 97.3                     & \textbf{95.3}                         \\ \hline
\end{tabular}}
\end{table}

The classification assessment measure obtained for both the scene classification systems shows that the proposed image decolorization method to perceptually improve the performance of the existing SVD image decolorization technique has slightly improved the overall accuracy of the AGMM-SVM scene classification system (81.8\% for NTSC , 81.6\% for SVD and 82.0\% for the proposed method (Modified SVD) tabulated in Table \ref{AGMM-SVM}) and has significantly improved the overall classification accuracy of the AGMM-SVM-DBN scene classification system (93.5\% for NTSC , 93.6\% for SVD and 95.3\% for the proposed method (Modified SVD) tabulated in Table \ref{AGMM-DBN-SVM}.

The class accuracy obtained using the proposed image decolorization method which are greater than that of the most commonly used NTSC rule and the existing SVD image decolorization for both the scene classification systems are marked as bold in Table \ref{AGMM-SVM} \& \ref{AGMM-DBN-SVM}. The improvement in the overall classification accuracy of the AGMM-SVM scene classification system by the proposed image decolorization method is due to the improvement in the class accuracy namely insidecity, mountain street and tall building (Table \ref{AGMM-SVM}) . Similarly, the significant improvement in the overall classification accuracy of the AGMM-DBN-SVM scene classification system by the proposed image decolorization method is due to the improvement in the classification accuracy of 6 classes out of available 8 classes (Table \ref{AGMM-DBN-SVM}).

The significant improvement in the AGMM-DBN-SVM scene classification system by the proposed image decolorization is due to the training of deep belief networks using the dense SIFT feature descriptors which extracted the significant chrominance information with the improved perceptual quality in the grayscale images converted using the proposed image decolorization method.

\subsection{Model level combination of the proposed image decolorization method and the most commonly used NTSC rule}

The analysis of classification accuracy obtained for each class shows the presence of complimentary class information provided by the proposed image decolorization method with respect to the standard NTSC rule. i.e., there are significant number of images misclassified by the standard NTSC rule are correctly classified by the proposed method for image decolorization and vice-versa. This motivated to combine the complimentary class information provided by the proposed image decolorization with the standard NTSC rule for the AGMM-DBN-SVM scene classification system whose, overall classification accuracy for all the image decolorization methods are greater than the AGMM-SVM image classification system.

Thus, the complimentary class information are combined based on the maximum probability score computed by the AGMM-DBN-SVM models for NTSC rule and the proposed method respectively. The class label associated with the model corresponding to the maximum probability score computed using the Eq. (\ref{probscore}) is selected as the class information of the input image.

\begin{equation}
\mathord{\buildrel{\lower3pt\hbox{$\scriptscriptstyle\frown$}}\over
 \theta } \, = \arg \max (P\,(x|\theta _1 \,)\,,\,P\,(x|\theta _2 \,)\,)\\
\label{probscore}
\end{equation}

where, $x$ represents the dense SIFT feature vectors of an image, $\theta _1$ and $\theta_2$ represents the AGMM-DBN-SVM models obtained using NTSC rule and the proposed method respectively.

\begin{table}[htbp]
\centering
\caption{Comparison of classwise accuracy (\%) and overall accuracy (\%) for the model level combination}
\label{modelcombination}
\vspace{0.5cm}
\scalebox{0.68}{
\begin{tabular}{|l|c|c|c|c|c|c|c|c|c|}
\hline
\multicolumn{1}{|c|}{\begin{tabular}[c]{@{}c@{}}Image Decolorization\\ Methods\end{tabular}} & \multicolumn{1}{l|}{opencountry} & \multicolumn{1}{l|}{coast} & \multicolumn{1}{l|}{forest} & \multicolumn{1}{l|}{highways} & \multicolumn{1}{l|}{insidecity} & \multicolumn{1}{l|}{mountain} & \multicolumn{1}{l|}{street} & \multicolumn{1}{l|}{tallbuilding} & \multicolumn{1}{l|}{overall accuracy} \\ \hline
NTSC rule                                                                                   & 90.9                             & 92.7                       & 92.1                        & 87.5                          & 93.3                           & 95.9                          & 96.4                        & 97.7                              & 93.5                                  \\ \hline
NTSC \& SVD                                                                                         & 91.3                             & 93.1                       & 94.7                        & 88.8                         & 93.8                            & 95.9                          & 95.8                        & 98.1                             & 94.1                                  \\ \hline
\begin{tabular}[c]{@{}l@{}}Proposed\\ (NTSC \& Modified SVD)\end{tabular}                                                                     & \textbf{91.9}                            & \textbf{94.6}                       & \textbf{95.2}                        & \textbf{91.3}                          &\textbf{95.2}                   & \textbf{96.7}                & \textbf{96.9}               & \textbf{98.1}                     & \textbf{95.0}                         \\ \hline
\end{tabular}}
\end{table}

The classification accuracy obtained for each class and the overall accuracy obtained for the AGMM-DBN-SVM model level combinations of the proposed image decolorization and the most commonly used NTSC rule is tabulated in Table \ref{modelcombination}. The classification accuracy of all the classes obtained using NTSC rule are improved by the above defined model level combination which led to the significant improvement in the overall classification accuracy of the NTSC rule from 93.5\% to 95.0\% (Table \ref{modelcombination}).

The overall classification accuracy of the AGMM-DBN-SVM system for NTSC rule is improved from 93.5\% to 94.1\% when the complimentary class information provided by the existing SVD image decolorization is combined with the NTSC rule. But, the improvement in the overall classification accuracy of the NTSC rule is greater (95\% (Table\ref{modelcombination})) when the complimentary class information provided by the proposed image decolorization method is combined with the NTSC rule. This is because of the improvement in the classification accuracy of all the classes due to the presence of complimentary class information captured by the proposed method of image decolorization. Thus, the effectiveness of the proposed method to perceptually improve the performance of the existing SVD image decolorization is also proved through the classification measures obtained for the model level combination system.

\section{Summary and Conclusion}
In the present work, a perceptually improved image decolorization technique using C2G-SSIM and SVD is proposed. The proposed method is experimented on two standard color-to-grayscale image conversion datasets namely Cadik and COLOR250. The effectiveness of the proposed image decolorization method is compared against 8 benchmark techniques based on the image quality assessment metric for image decolorization called C2G-SSIM, success rate.

The experimental results analysis based on C2G-SSIM and success rate shows that the proposed color-to-grayscale image conversion algorithm performs better than all the existing benchmark techniques for image decolorization. Also, the role of incorporating significant chrominance information by the proposed image decolorization method for scene classification is confirmed by the improvement in the scene classification accuracy obtained for two different scene classification systems called AGMM-SVM and AGMM-DBN-SVM developed using dense scale invariant feature transform (SIFT) as features.

Also, a new method is devised to combine the models obtained using the proposed image decolorization method with the standard color-to-grayscale image conversion technique which improved the overall performance of the scene classification system. The overall scene classification accuracy obtained using the newly devised model level combination of the proposed image decolorization and the standard color-to-grayscale image conversion is better than the classification accuracy obtained using the standard image decolorization method independently.

The main contributions of the present work are as follows:

\begin{itemize}
\item Developed a new algorithm to improve the perceptual quality of the image decolorization.
\item Significant improvement in the scene classification system using the dense SIFT features extracted from the grayscale images converted by the proposed image decolorization method as input to the AGMM-DBN-SVM classifier.
\item The combination of the AGMM-DBN-SVM models of the proposed image decolorization technique with the most commonly used NTSC rule for image conversion improved the scene classification performance of the existing standard NTSC method for image decolorization.
\end{itemize}

As the future scope of the present work, the performance of scene classification by the proposed image decolorization method can be analyzed at feature level using the different existing benchmark feature extraction techniques for scene classification.

\bibliographystyle{plain}
\bibliography{refer}

\end{document}